\documentclass[runningheads]{llncs}
\usepackage{graphicx}

\usepackage{amssymb}
\usepackage{amsmath}
\usepackage{color}
\usepackage{algpseudocode}
\usepackage{soul}
\usepackage{algorithm}% http://ctan.org/pkg/algorithm
\usepackage[T1]{fontenc}
\usepackage{url}
\usepackage{wrapfig,lipsum,booktabs}
\usepackage{hyperref}

\begin{document}

\title{DECAPS: Detail-Oriented Capsule Networks}

\author{Aryan Mobiny \and Pengyu Yuan \and Pietro Antonio Cicalese \and
Hien Van Nguyen}
%
%index{Mobiny, Aryan} 
%index{Yuan, Pengyu}
%index{Cicalese, Pietro Antonio} 
%index{Van Nguyen, Hien} 
%
\authorrunning{A. Mobiny et al.}
\institute{Department of Electrical and Computer Engineering \\ University of Houston \\
\email{amobiny@uh.edu}}
%

% \author{Anonymous}
% \institute{Anonymous Organization
% \email{***@***.***}}

\maketitle              % typeset the header of the contribution

\begin{abstract}
% \vspace{-6mm}
Capsule Networks (CapsNets) have demonstrated to be a promising alternative to Convolutional Neural Networks (CNNs). However, they often fall short of state-of-the-art accuracies on large-scale high-dimensional datasets. We propose a Detail-Oriented Capsule Network (DECAPS) that combines the strength of CapsNets with several novel techniques to boost its classification accuracies. First, DECAPS uses an Inverted Dynamic Routing (IDR) mechanism to group lower-level capsules into heads before sending them to higher-level capsules. This strategy enables capsules to selectively attend to small but informative details within the data which may be lost during pooling operations in CNNs. Second, DECAPS employs a Peekaboo training procedure, which encourages the network to focus on fine-grained information through a second-level attention scheme. Finally, the distillation process improves the robustness of DECAPS by averaging over the original and attended image region predictions. We provide extensive experiments on the CheXpert and RSNA Pneumonia datasets to validate the effectiveness of DECAPS. Our networks achieve state-of-the-art accuracies not only in classification (increasing the average area under ROC curves from 87.24\% to 92.82\% on the CheXpert dataset) but also in the weakly-supervised localization of diseased areas (increasing average precision from 41.7\% to 80\% for the RSNA Pneumonia detection dataset). 

% Our approach serves as a supplementary module that can be added to any network designed for object localization.

% These properties are especially useful for medical image diagnosis where small details can completely change the diagnostic outcome.

%  We demonstrate that the DECAPS architecture can intelligently attend to important details in data and achieve higher accuracies on large-scale, high-dimensional datasets.

% \vspace{-3mm}
% \vspace{-3mm}

\end{abstract}
\section{Introduction}
% \vspace{-2mm}
Convolutional neural networks (CNNs) have achieved state-of-the-art performance in many computer vision tasks due to their ability to capture complex representations of the desired target concept \cite{he2016deep,krizhevsky2012imagenet,ravindran2019assaying,ren2015faster,mobiny2019risk,shahraki2018graph}. These architectures are composed of a sequence of convolutional and pooling layers, with max-pooling being popularized in the literature due to its positive effect on performance. The max-pooling operation allows CNNs to achieve some translational invariance (meaning they can identify the existence of entities regardless of their spatial location) by attending only to the most active neuron. However, this operation has been criticized for destroying spatial information which can be valuable for classification purposes. 
%The max-pooling operation has, however, been criticized for its conceptual deficiency; it fails to comprehensively route information as it only attends to the most active neuron. This allows CNNs to achieve some translational invariance (meaning they can identify the existence of entities regardless of their spatial location), but destroys location and pose information, which can be valuable for classification purposes.Capsule networks (CapsNets) were introduced as an alternative architecture to CNNs, and are meant to address these fundamental deficiencies 
Capsule networks (CapsNets) aim to address this fundamental problem of max pooling and has become a promising alternative to CNNs \cite{sabour2017dynamic}. Previous works have demonstrated that CapsNets possess multiple desirable properties; they are able to generalize with fewer training examples, and are significantly more robust to adversarial attacks and noisy artifacts \cite{mobiny2018fast,hinton2018matrix}. CapsNets utilize view-point invariant transformations that learn to encode the relative relationships (including location, scale, and orientation) between sub-components and the whole object, using a dynamic routing mechanism to determine how information should be categorized. This effectively gives CapsNets the ability to produce interpretable hierarchical parsing of each desired scene. By looking at the paths of the activations, we can navigate the hierarchy of the parts and know exactly the parts of an object. This property has prompted several research groups to develop new capsule designs and routing algorithms \cite{hinton2018matrix,ahmed2019star,mobiny2019automated,kosiorek2019stacked}. 

In recent years, CapsNets have been widely adopted and used in various medical image analysis tasks \cite{afshar20203d,iesmantas2018convolutional,jimenez2018capsule}. Jiao \emph{et al.} used CapsNet for the diagnosis of mild cognitive impairment \cite{jiao2019dynamic}. Mobiny \emph{et al.} proposed Fast CapsNet which exploits novel techniques to improve the inference time and prediction performance of CapsNets for the lung nodule classification in 3D CT scans \cite{mobiny2018fast}. Lalonde \emph{et al.} proposed SegCaps to expand capsule networks to segment pathological lungs from low dose CT scans \cite{lalonde2018capsules}.

In medical image analysis, identifying the affected area and attending to small details is critical to diagnostic accuracy. In this paper, we propose a novel version of CapsNets called Detail-Oriented Capsule Networks (DECAPS) which simulate this process by attending to details within areas that are relevant to a given task while suppressing noisy information outside of the region of interest (or ROI). We can effectively describe our architecture as having both a coarse and fine-grained stage. First, the architecture groups capsules into submodules named capsule heads, each of which is trained to extract particular visual features from the input image. It then employs an inverted routing mechanism in which groups of lower-level capsules compete with each other to characterize the task-specific regions in the image, generating coarse level predictions. The ROIs are then cropped and used in the fine-grained prediction scheme, where the model learns to interpret high-resolution representations of areas of interest. The two predictions are then combined and generate a detail-oriented output which improves performance. We will make DECAPS implementation publicly available.
%at \url{https://github.com/hula-ai/DECAPS}.

% enerate capsule-head activation maps (CHAMs) to describe the role of each capsule for a given output prediction. To focus on the ROI and capture detailed information, we use the CHAMs to generate an input for the second-level attention scheme, where the classifier learns to interpret high resolution representations of areas of interest. By combining this information with the coarse level, we effectively generate a 

% Combining the multi-head structure and our proposed routing mechanism allows the model to aggregate information from different visual feature subspaces from different locations in the image.

% an be further employed to extract fine-grained information and localize the ROIs.

% \vspace{-2mm}
\section{Background on Capsule Networks} 
% \vspace{-4mm}
A CapsNet is composed of a cascade of capsule layers, each of which contains multiple capsules. A capsule is the basic unit of CapsNets and is defined as a \emph{group of neurons} whose output forms a \textit{pose} vector. This is in contrast to traditional deep networks which use neurons as their basic unit. Let $\Omega_L$ denote the sets of capsules in layer $L$. Each capsule $i \in \Omega_L$ has a pose vector $\text{p}^L_i$. The length (the norm or magnitude) of the pose vector encodes the probability that an object of interest is present, while its direction represents the object's pose information, such as location, size, and orientation. The $i$-th capsule in $\Omega_L$ propagates its information to $j$-th capsule in $\Omega_{L+1}$ through a linear transformation $\text{v}^L_{ij} = \text{W}^L_{ij} \text{p}^L_i$, where $\text{v}^L_{ij}$ is called a \emph{vote} vector. The pose vector of capsule $j \in \Omega_{L+1}$ is a convex combination of all the votes from child capsules: $\text{p}^{(L+1)}_j=\sum_i r_{ij} \text{v}^L_{ij} $, where $r_{ij}$ are routing coefficients and $\sum_{i} r_{ij}=1$. These coefficients are determined by the dynamic routing algorithm \cite{sabour2017dynamic} which iteratively increases the routing coefficients $r_{ij}$ if the corresponding voting vector $\text{v}_{ij}^L$ is similar to $\text{p}_j^{L+1}$ and vice versa. Dynamic routing ensures that the output of each child capsule gets sent to proper parent capsules. Through this process, the network gradually constructs a transformation matrix for each capsule pair to encode the corresponding part-whole relationship and retains geometric information of the input data.

\noindent \textbf{Notations:} Throughout the paper, $r$, $\text{r}$, $\text{R}$, $\mathbf{R}$ represent a scalar, a vector, a 2D matrix, and a tensor (i.e. a higher dimensional matrix; usually a 3D matrix of capsule activations), respectively. Note that multiplying a transformation matrix and a tensor of poses is equivalent to applying the transformation to each pose.

%The relationship between the $i$-th capsule in $\Omega_L$ and the $j$-th capsule in $\Omega_{L+1}$ is encoded using a linear transformation matrix $\text{W}^L_{ij}$. The information is propagated through the network following  $\text{v}^L_{ij} = \text{W}^L_{ij} \text{p}^L_i$, in which the \textit{vote} vector $\text{v}^L_{ij}$ represents the characterization of the $j$-th capsule in $\Omega_L$ by the $i$-th capsule of $\Omega_L$.
% Through this process, the network gradually constructs a transformation matrix for each capsule pair to encode the corresponding part-whole relationship. After computing the prediction vectors (or votes), the lower-level capsules route their information to the parent capsules that agree most with their predictions. Dynamic Routing is the mechanism that ensures that the output of each child capsule gets sent to the proper parent capsules. The pose vector of capsule $j \in \Omega_{L+1}$ is a linear combination of the votes: $\text{p}^{(L+1)}_j=\sum_i r_{ij} \text{v}^L_{ij} $, where $r_{ij}$ is a routing coefficient determined by the routing algorithm. The routing procedure ensures that the predictions that are far from the general consensus have less impact on the output of each layer. 

% In other words, the process 

% \vspace{-2mm}
\section{Detail-Oriented Capsule Network} % \vspace{-3mm}
\begin{figure}[!t]
\centering
\includegraphics[width=\textwidth]{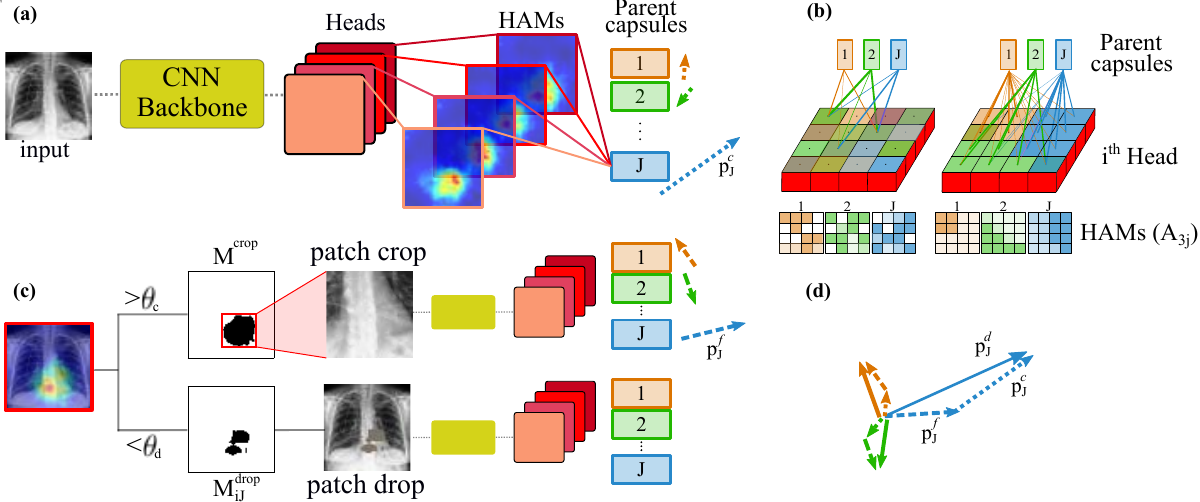}
% \vspace{-2mm}
\caption{\textbf{(a)}: DECAPS architecture. Head activation maps (HAMs) are presented for the $J^{th}$ class. \textbf{(b)}: Dynamic (left) vs. Inverted dynamic routing (right). Inverted dynamic routing places the competition between children capsules of a head yielding discriminative and localized HAMs. \textbf{(c)}: Peekaboo training. \textbf{(d)}: The distillation process to fine-tune the coarse-grained prediction ($\text{p}^c$) using the fine-grained prediction ($\text{p}^f$). }
%In dynamic routing, child capsules rout independent of one another; resulting in maps with activations spread all over the place. } 
% \vspace{-5mm}
\label{model}
\end{figure}

In the original CapsNet \cite{sabour2017dynamic}, the vote of each child capsule contributes directly to the pose of all parent capsules. This ultimately has a negative effect on the quality of the final prediction, due to the noisy votes derived from non-descriptive areas. DECAPS utilize a modified architecture, loss, and routing mechanism  that favors votes from ROIs, thus improving the quality of the inputs being routed to the parent capsules. Inspired by the Transformers architecture \cite{vaswani2017attention}, we group capsules within a grid and call them \textit{Capsule Heads} (see Fig. \ref{model} (a)). One can think of a capsule head as being a grid of capsules that routes information independently of the other heads. Ideally, each capsule head is responsible for detecting a particular visual feature in the input image. To accomplish this, each head shares a transformation matrix $\text{W}^L_{ij}$ between all capsules for each output class. This contrasts against the original architecture which uses one transformation matrix per capsule, per class. This reduces the required number of trainable parameters by an order of head size (i.e. the number of capsules within a head, $26 \times 26$ in our proposed DECAPS); this allows DECAPS to properly scale to large, high-dimensional input images. Additionally, we use the head activation regularization loss (explained in section 3.2) to force the capsules within a head to seek the same semantic concept for each diagnostic task.

Let $\mathbf{P}^L_i \in \mathbb{R}^{h_L \times w_L \times d_L}$ denote the pose matrix of the capsules of the $i^{th}$ head where $h_L$ and $w_L$ represent the height and width of the head respectively, and $d_L$ is the capsule dimension (i.e. the number of hidden units grouped together to make capsules in layer $L$). Note that $i$ is the child capsule index in the original CapsNet, but is changed to represent the head index in our architecture. We want the length of the pose vector of each capsule to represent the probability of existence for a given entity of interest in the current input. The capsule outputs are passed through a nonlinear squash function \cite{sabour2017dynamic} to ensure that the length of the pose vectors is normalized between zero and one. Then we say that $\mathbf{V}^L_{ij}=\text{W}^L_{ij}\mathbf{P}^L_i$ is the votes from the capsules of the $i^{th}$ head to the $j^{th}$ parent capsule. To preserve the capsule's location, we perform Coordinate Addition: at each position within a capsule head, the capsule's relative coordinates (row and column) are added to the final two entries of the vote vector \cite{hinton2018matrix}. Once we have generated the votes, the routing mechanism determines how information should flow to generate each parent's pose vector.

% proposed routing procedure determines the information flow from children to parent capsules.

% The constraint is that the routing coefficients from a child capsule to all parent capsules sum to 1 ($\sum_j r_{ij}=1$). 

% filter out the noisy information in a way that capsules of the subsequent layer receive a cleaner input signal only from the capsules within the ROI to determine the pose of the objects more accurately. 

% \begin{wrapfigure}{r}{0.5\textwidth}
% \vspace{-15mm}
%   \begin{center}
% \includegraphics[width=0.45\textwidth]{figs/inverted_routing.pdf}
%   \end{center} 
%   \vspace{-7mm}
% \caption{Dynamic (left) vs. Inverted dynamic routing (right). In dynamic routing, child capsules rout independent of one another; resulting in maps with activations all over the place. Inverted routing, however, places the competition between children capsules of a head yielding discriminative and localized class-specific attention maps}
% \label{fig:idr}
%   \vspace{-5mm}
% \end{wrapfigure}

\begin{algorithm}[!t]
\caption{Inverted Dynamic Routing (IDR). Note that $i$ and $j$ are the indices of capsule heads in layer $L$ and $L+1$ respectively.}
\label{algorithm}
\begin{algorithmic}[1]
\Procedure{IDR}{$\mathbf{V}^{L}_{ij}, n_{\text{iter}}$}\Comment{given the votes and number of routing iterations}
\State $\text{R}_{ij}^{\text{pre}}\gets 0, \quad \forall j$ \Comment{initialize the routing coefficients}
\For{$n_{\text{iter}}$ iterations}
% \State $\mathbf{C}_{ij}\gets \underset{i}{ \text{{\fontfamily{lmtt}\selectfont softmax}} } (\mathbf{C}_{ij}^{\text{pre}})$ \Comment{softmax-normalized coupling coefficients}
\State $\text{R}_{ij}\gets \text{{\fontfamily{lmtt}\selectfont softmax}} (\text{R}_{ij}^{\text{pre}})$ \Comment{softmax among capsules in head $i$}

\State $\mathbf{\Tilde{A}}_{ij} \gets \text{R}_{ij} \odot \mathbf{V}^{L}_{ij}$
\Comment{$\odot$ is the Hadamard product}

\State $\text{p}_j^{L+1} \gets \text{{\fontfamily{lmtt}\selectfont squash}}(\sum_i \sum_{xy} \mathbf{\Tilde{A}}_{ij})$
\vspace{0.5mm} \Comment{$\sum_{xy}$ is the sum over spatial locations}

\State $\text{R}_{ij}^{\text{pre}} \gets \text{R}_{ij}^{\text{pre}} + \text{p}_j^{(L+1)} .  \mathbf{V}^{L}_{ij}$
\EndFor

\State $\text{A}_{ij} \gets \text{{\fontfamily{lmtt}\selectfont length}} (\mathbf{\Tilde{A}}_{ij})$
\Comment{$\text{{\fontfamily{lmtt}\selectfont length}}$ computes Eq. \eqref{eq1}}
\State \textbf{return} $\text{p}_j^{L+1}, \text{A}_{ij}$
\EndProcedure
\end{algorithmic}
\end{algorithm}

% \vspace{-3mm}
\subsection{Inverted Dynamic Routing}
% \vspace{-3mm}
Dynamic routing \cite{sabour2017dynamic} is a \textit{bottom-up} approach which forces higher-level capsules to compete with each other to collect lower-level capsule votes. We propose an inverted dynamic routing (IDR) technique which implements a \textit{top-down} approach, effectively forcing lower-level capsules to compete for the attention of higher-level capsules (see Fig. \ref{model} (b)). During each iteration of the routing procedure, we use a softmax function to force the routing coefficients between all capsules of a single head and a single parent capsule to sum to one (see Algorithm \ref{algorithm}). The pose of the $j^{th}$ parent capsule, $\text{p}^{L+1}_j$, is then set to the squashed weighted-sum over all votes from the earlier layer (line 6 in Algorithm \ref{algorithm}). Given the vote map computed as $\mathbf{V}^L_{ij}=\text{W}^L_{ij}\mathbf{P}^L_i \in \mathbb{R}^{h_L\times w_L\times d_{L+1}}$, the proposed algorithm generates a routing map $\text{R}_{ij} \in \mathbb{R}^{h_L\times w_L}$ from each capsule head to each output class. The voting map describes the children capsules' votes for the parent capsule's pose. The routing map depicts the weights of the children capsules according to their agreements with parent capsules, with winners having the highest $r_{ij}$. We combine these maps to generate head activation maps (or HAMs) following

% \vspace{-2mm}
\begin{equation}
\label{eq1}
\small{\text{A}_{ij} = ({\sum\nolimits_{d} \mathbf{\Tilde{A}}_{ij}^2})^{1/2}, \quad \text{where} \quad \mathbf{\Tilde{A}}_{ij} = \text{R}_{ij} \odot \mathbf{V}^{L}_{ij}}
\end{equation}

% \vspace{-1mm}

\noindent
where $\text{A}_{ij}$ is the HAM from the $i^{th}$ head to the $j^{th}$ parent, and $\sum_d$ is the sum over $d_{L+1}$ channels along the third dimension of $\mathbf{V}_{ij}^L$.  $\text{A}_{ij}$ highlights the informative regions within an input image corresponding to the $j^{th}$ class, captured by the $i^{th}$ head. IDR returns as many activation maps as the number of capsule heads per output class (see Fig. \ref{model} (a)). Class-specific activation maps are the natural output of the proposed framework, unlike CNNs which require the use of additional modules, such as channel grouping, to cluster spatially-correlated patterns \cite{zheng2017learning}. We utilize the activation maps to generate ROIs when an object is detected. This effectively yields a model capable of weakly-supervised localization which is trained end-to-end; we train on the images with categorical annotations and predict both the category and the \emph{location} (i.e. mask or bounding box) for each test image. This framework is thus able to simultaneously generate multiple ROIs within the same image that are associated with different medical conditions.

% \vspace{-4mm}
\subsection{Loss Function}
% \vspace{-2mm}

% Our loss consists of two terms: Margin loss \cite{sabour2017dynamic} for the classification and a regularization term to manipulate the attention maps.
The loss function we define is the sum of two terms and is described as follows:

\vspace{0.5mm}
\noindent
\textbf{Margin Loss:} We use margin loss to enforce the activation vectors of the top-level capsule $j$ to have a large magnitude if and only if the object of the corresponding class exists in the image \cite{sabour2017dynamic}. The total margin loss is the sum of the losses for each output capsule as given by

% \vspace{-5mm}
\begin{equation}
\label{eq2}
\small{ L_{\text{margin}}=\sum\nolimits_j [ T_j \; \text{max}(0, m^+ - \|\text{p}_j\|)^2 + \lambda (1-T_j) \; \text{max}(0, \|\text{p}_j\|-m^-)^2 ] }
\end{equation}

% \vspace{-3mm}

\noindent where $T_j = 1$ when class $j$ is present (else $T_j = 0$). Minimizing this loss forces $\|\text{p}_j\|$ of the correct class to be higher than $m^+$, and those of the wrong classes to be lower than $m^-$. In our experiments, we set $m^+ = 0.9$, $m^- = 0.1$, and $\lambda = 0.5$.

\vspace{0.5mm}
\noindent
\textbf{Head Activation Regularization:}
% Intuitively, maximizing the inter-class difference while minimizing the intra-class variations helps improving the discriminative power of the learned deep features. 
We propose a regularization loss function to supervise the head activation learning process. We want each head activation map $\text{A}_{ij}$ to capture a unique semantic concept of the $j^{th}$ output category. Inspired by center loss \cite{wen2016discriminative}, we define a feature template $\text{t}_{ij} \in \mathbb{R}^{d_{L+1}}$ for the $i^{th}$ semantic concept of the $j^{th}$ output category. We compute the semantic features $\text{f}_{ij} \in \mathbb{R}^{d_{L+1}}$ using the information routed from the $i^{th}$ head to the $j^{th}$ output category. While the magnitude of $\text{f}_{ij}$ represents the presence of the desired semantic concept, the orientation captures the instantiation parameters (i.e. pose, deformation, texture, etc.). We, therefore, want to regularize the orientation of $\text{f}_{ij}$ for a given capsule head to guarantee that it is capturing the same semantic concept among all training images. Each value of $\text{t}_{ij}$ is initialized to zero and is updated using a moving average as

% \vspace{-2.2mm}
\begin{equation}
% \vspace{-2.5mm}
\label{eq4}
\small{\text{t}_{ij} \gets \text{t}_{ij} + \gamma (\hat{\text{f}}_{ij} - \hat{\text{t}}_{ij}), \quad \text{where} \; \text{f}_{ij}=1/n_i\small{\sum\nolimits}_{xy}\mathbf{\Tilde{A}}_{ij}, \; 
% \hat{\text{f}}_{ij} = \text{f}_{ij}/ \lVert \text{f}_{ij} \rVert
}
\end{equation}

\noindent
where $\hat{\text{f}}_{ij}$ and $\hat{\text{t}}_{ij}$ are the normalized vectors, $\gamma$ is the update step, which we set to $10^{-4}$, while $n_i$ represents the number of capsules in head $i$. To accomplish this, we penalize the network when the orientation of a head's features $\text{f}_{ij}$ deviate from the template $\text{t}_{ij}$ following

% \vspace{-1mm}

\begin{equation}
\label{eq3}
% \vspace{-1mm}
\small{L_{\text{HAR}}=\frac{1}{IJ}\sum\nolimits_i\sum\nolimits_j  (1- \text{{\fontfamily{lmtt}\selectfont cosine}}(\text{f}_{ij}, \text{t}_{ij}) )}
% = \frac{\text{f}_{ij} \odot \text{t}_{ij}}{\lVert\text{f}_{ij} \rVert \lVert \text{t}_{ij} \rVert}
\end{equation}

\noindent
where $I$ and $J$ represent the total number of child and parent capsules. %Note that $\text{{\fontfamily{lmtt}\selectfont cosine}}$ is the cosine similarity function that computes the similarity between the two vectors according to their orientation, and not their magnitude.

% \vspace{-2mm}
\subsection{Peekaboo: the activation-guided training}
% \vspace{-1mm}
To further promote DECAPS to focus on fine-grained details, we propose the Peekaboo strategy for capsule networks. Our strategy boosts the performance of DECAPS by forcing the network to look at all relevant parts for a given category, not just the most discriminative parts \cite{singh2017hide}. Instead of hiding random image patches, we use the HAMs to guide the network's attention process. For each training image, we randomly select an activation map $\text{A}_{ij}$ for each recognized category. Each map is then normalized in the range $[0, 1]$ to get the normalized HAM $\text{A}^{*}_{ij} \in \mathbb{R}^{h_L\times w_L}$. We then enter a two step process: patch cropping, which extracts a fine-grained representation of the ROI to learn how to encode details, and patch dropping, which encourages the network to attend to multiple ROIs. In patch cropping, a mask $\text{M}^{crop}_{ij} \in \mathbb{R}^{h_L\times w_L}$ is obtained by setting all elements of $\text{A}^{*}_{ij}$ which are less than a cropping threshold $\theta_c \in [0, 1]$ to 0, and 1 otherwise. We then find the smallest bounding box which covers the entire ROI, and crop it from the raw image (Fig. \ref{model} (c)). It is then upsampled and fed into the network to generate a detailed description of the ROI. During the patch dropping procedure, $\text{M}^{drop}_{ij}$ is used to remove the ROI from the raw image by using a dropping threshold $\theta_d \in [0, 1]$. The new patch-dropped image is then fed to the network for prediction. This encourages the network to train capsule heads to attend to multiple discriminative semantic patterns. At test time, we first input the whole image to obtain the coarse prediction vectors ($\text{p}^c_j$ for the $j^{th}$ class) and the HAMs $\text{A}_{ij}$ from all capsule heads. We then average all maps across the heads, crop and upsample the ROIs, and feed the regions to the network to obtain the fine-grained prediction vectors ($\text{p}^f_j$). The final prediction $\text{p}^d_j$, referred to as distillation (Fig. \ref{model} (d)), is the average of the $p_j^c$ and $p_j^f$.

% \vspace{-2mm}
\section{Experiments and Results}
% \vspace{-2mm}
\noindent
\textbf{Implementation details:} In our experiments, we use Inception-v3  as the backbone and take the \textit{Mix6e} layer output as the CNN feature maps. We then compress the feature maps using $1\times 1$ convolutional kernels to generate 256 maps. We split the maps into four capsule heads, each of which includes a grid of 64-dimensional capsules. These heads employ the described inverted routing procedure to route into the final 16-dimensional class capsules. The best performance was achieved using 3 routing iterations, $\theta_c=0.5$, and $\theta_d=0.3$. The network is trained using the Adam optimizer with $\beta_1=0.5$, $\beta_2=0.999$ and a learning rate of $10^{-4}$ which is fixed for the duration of training. 
% Images are fed into the network with size $448\times 448$ pixels (yielding a $26 \times 26$ capsule map per head) while batches are sampled using a fixed batch size of 16.

% \vspace{1mm}
\noindent
\textbf{Datasets:} We use two datasets to evaluate the performance of the proposed DECAPS architecture. The CheXpert \cite{irvin2019chexpert} radiography dataset is used for the detection of 5 selected pathologies, namely Atelectasis, Cardiomegaly, Consolidation, Edema, and Pleural effusion (see Table 1 of \cite{irvin2019chexpert} for more information on the data distribution). 
%We also include a ``No Finding'' category for images that are diagnosed with none of the 5 pathologies. Our observations show that this change helps lower-level capsules to vote for the parent capsule associated to the No Finding category for the normal cases, and boosts the prediction performance of the model. 
We also report our results on the RSNA Pneumonia detection data which includes bounding box annotations of the affected regions in the images \cite{RSNA}. It is important to note that our approach only uses the category labels (not the bounding boxes) for the localization of pneumonia localization.

% \vspace{1mm}
\noindent
\textbf{Evaluation metrics:} We use the area under ROC curve (AUC) to report the prediction accuracy on the CheXpert dataset. We use the mean intersection over union (mIoU) to evaluate the localization accuracy of the model on the RSNA dataset. We also compute the average precision (AP) at different IoU thresholds. At each threshold, a true positive (TP) is counted when a predicted object matches a ground truth object with an IoU above the threshold. A false positive (FP) indicates a predicted object had no associated ground truth object. A false negative (FN) indicates a ground truth object had no associated predicted object. We then calculated AP as $\text{TP}/(\text{TP+FP+FN})$ over all test samples \cite{RSNA}.

\begin{table}[!b]
\centering
\caption{Prediction performance of models trained on the CheXpert dataset. For each model, average result is reported over the best 10 trained model checkpoints.}
% \vspace{-3mm}
\resizebox{\columnwidth}{!}{
\begin{tabular}{lcccccc}
\hline
  & \textbf{Cardiomeg.} & \textbf{Edema} & \textbf{Consolid.} & \textbf{Atelectasis} & \textbf{Pleural Eff.} & \textbf{mAUC (\%)}\\ \hline
 Inception-v3 \cite{szegedy2016rethinking} & \;$0.841(\pm0.052)$ & \;$0.876(\pm0.055)$ & \;$0.891(\pm0.044)$  & \;$0.833(\pm0.032)$ & \;$0.921(\pm0.038)$ & \;$87.24$\;\\
DenseNet121 \cite{irvin2019chexpert} & $0.832(\pm0.047)$ & $0.941(\pm0.031)$  & $0.899(\pm0.037)$ & $0.858(\pm0.042)$ & $0.934(\pm0.027)$  & $89.28$\\
DenseNet121+HaS \cite{singh2017hide} \;\;  & $0.849(\pm0.041)$  & $0.940(\pm0.055)$ & $0.904(\pm0.039)$  & $0.867(\pm0.050)$ & $0.938(\pm0.024)$ & $89.96$\\ 
CapsNet \cite{sabour2017dynamic} \;\;  & $0.835(\pm0.033)$  & $0.915(\pm0.038)$ & $0.890(\pm0.035)$  & $0.845(\pm0.031)$ & $0.949(\pm0.033)$ & $88.68$\\ \hline
\textbf{DECAPS} & $0.852(\pm0.048)$ & $0.935(\pm0.039)$ & $0.897(\pm0.028)$  & $0.865(\pm0.045)$ & $0.946(\pm0.022)$  & $89.90$\\
\textbf{DECAPS+Peekaboo} \; & $\mathbf{0.895(\pm0.044)}$ & $\mathbf{0.972(\pm0.027)}$  & $\mathbf{0.913(\pm0.033)}$ & $\mathbf{0.883(\pm0.029)}$ & $\mathbf{0.978(\pm0.019)}$  & $\mathbf{92.82}$\\ \hline
\end{tabular}
}
\label{table:1}
\end{table}

\begin{figure}[!t]
\centering
\includegraphics[width=\textwidth]{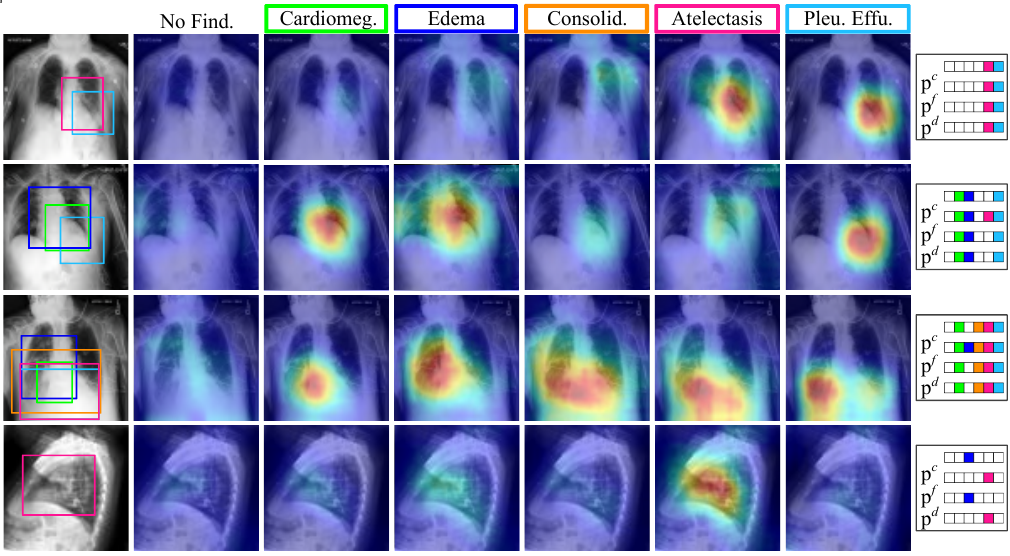}
% \vspace{-3mm}
\caption{Qualitative results on the CheXpert dataset. 
%Each row includes a sample input with bounding boxes of predicted pathologies (left column), and average HAMs (over all capsule heads) per category.
True pathologies (T), coarse ($\text{p}^c$), fine-grained ($\text{p}^f$), and distilled ($\text{p}^d$) predictions are presented for each case.} 
\label{fig1}
% \vspace{-5mm}
\end{figure}

\noindent
\textbf{Results on CheXpert Dataset:} 
The quantitative classification results are summarized in Table \ref{table:1}. We compare our results with Inception-v3 (which is the backbone used in our framework), DenseNet121 (the best performing baseline CNN according to \cite{irvin2019chexpert}), a DenseNet121 model trained with the Hide-and-Seek (HaS) strategy \cite{singh2017hide} to boost the weakly-supervised localization of the model, and the vanilla CapsNet with the same backbone. The vanilla DECAPS architecture yielded significantly higher classification accuracies than the baseline networks and achieves performance on par with DenseNet121+HaS. Adding the proposed Peekaboo method to our framework significantly improves the prediction and localization performance of the model. Examples are shown in Fig. \ref{fig1}. Each HAM is activated when the model detects a visual representation associated with the pathology of interest.  It highlights the ROI which will be cropped and passed through the fine-grained prediction stage to then generate the distilled prediction. The first row in Fig. \ref{fig1} shows an example with accurate classifications, while the second and third rows show samples that benefit from fine-grained prediction and distillation (atelectasis and edema are correctly removed). The fourth row shows a failure case that is diagnosed as Edema, but predicted as Atelectasis (see more examples and ablation study in the supplementary material section). 
% We performed an ablation study on our model trained on the CheXpert dataset to demonstrate that each component (namely IDR, patch crop and drop, and distillation) is effective in improving model performance, as shown in Table \ref{table:2}.

% \vspace{-5mm}
\begin{table}[!b]
\centering
\caption{Test prediction accuracy (\%), mean intersection over union (mIoU), and average precision (AP) over various IoU thresholds for RSNA Pneumonia detection.}
% \vspace{-3mm}
\resizebox{0.98\columnwidth}{!}{
\begin{tabular}{lccccccc}
\hline
 & \;Unsupervised\; & \;\;\%Acc\;\; & mIoU & AP$_{0.3}$ & AP$_{0.4}$ & AP$_{0.5}$ & AP$_{0.6}$ \\ \hline
Inception-v3+HaS & \checkmark & 87.14 & 0.314($\pm$0.321) & 0.417 & 0.370 & 0.241 & 0.194 \\
Faster-RCNN &  & 92.77 & \textbf{0.611($\pm$0.125)} & \textbf{0.887} & \textbf{0.853} & \textbf{0.718} & \textbf{0.561} \\ \hline
\textbf{DECAPS (level-1)} & \checkmark & \;86.25\; & \;0.401($\pm$0.176)\; & \;0.642\; & \;0.537\; & \;0.460\; &  \;0.322\; \\
\textbf{DECAPS (level-2)} & \checkmark & \textbf{94.02} & 0.509($\pm$0.130) & 0.800 & 0.771 & 0.594 & 0.471    \\ \hline

\end{tabular}
}
\label{table:3}
\end{table}

\begin{figure}[!t]
\centering
\includegraphics[width=\textwidth]{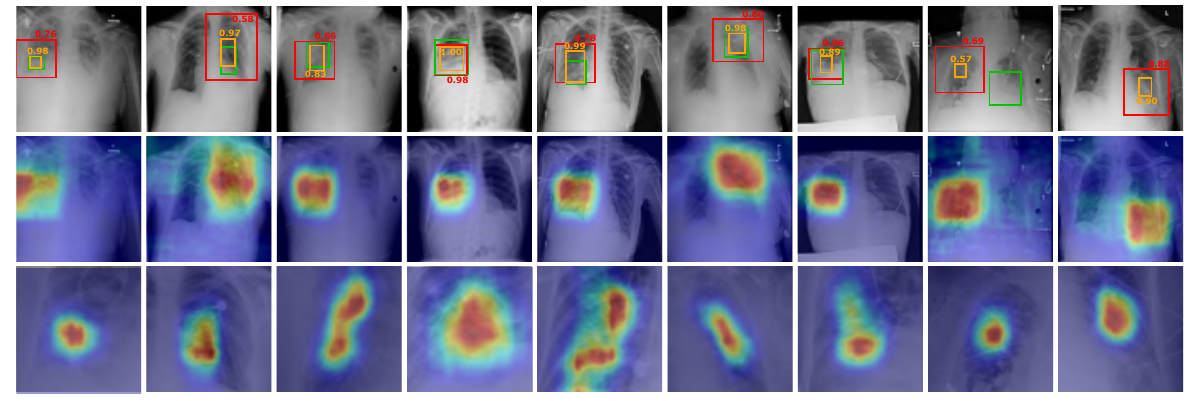}
% \vspace{-3mm}
\caption{Qualitative results on the RSNA dataset. \textbf{Top:} Input images with ground truth (green), level-1 (red), and level-2 (orange) bounding boxes. \textbf{Middle} Course-grained (level-1) activation maps. \textbf{Bottom:} Fine-grained (level-2) activation maps.}
\label{fig2}
% \vspace{-6mm}
\end{figure}

\vspace{1mm}
\noindent
\textbf{Results on RSNA Dataset:} We compare our qualitative results with a weakly-supervised localization approach \cite{singh2017hide}, as well as a supervised (Faster RCNN \cite{ren2015faster}) detection method as shown in Table \ref{table:3}. The prediction and localization metrics are computed at two levels for the DECAPS model: level-1 refers to the coarse prediction on the whole image while level-2 refers to the localization result of the fine-grained prediction (i.e. the ROI within the cropped region, examples shown in Fig. \ref{fig2}). We observe that the fine-grained prediction stage significantly improves the weakly-supervised localization performance over the coarse prediction stage and the baseline weakly supervised method (Inception-v3+HaS). We also note that the prediction accuracy of the fine-grained prediction stage exceeds the supervised Faster-RCNN prediction diagnosis, while localization accuracy is lower. We hypothesize that this is due to the coarse nature of the ground truth bounding boxes which also capture superfluous information from other tissues.

% \vspace{-3mm}
\section{Conclusions}
% \vspace{-4mm}
In this work, we present a novel network architecture, called DECAPS, that combines the strength of CapsNets with detail-oriented mechanisms. DECAPS is first applied to the whole image to extract global context and generate saliency maps that provide coarse localization of possible findings. This is analogous to a radiologist roughly scanning through the entire image to obtain a holistic view. It then focuses in the informative regions to extract fine-grained visual details from the ROIs. Finally, it employs a distillation process that aggregates information from both global context and local details to generate the final prediction. DECAPS achieves the highest accuracies on CheXpert and RSNA Pneumonia datasets. Despite being trained with only image-level labels, DECAPS are able to accurately localize the region of interests which enhances the model's interpretability. We expect our method to be widely applicable to image detection and recognition tasks, especially for medical image analysis tasks where small details significantly change the diagnostic outcomes. %Future studies should have a radiologist compare the ground truth and the generated ROIs as a means of quality assessment.

% 

% \begin{figure}[!t]
% \centering
% \includegraphics[width=0.9\textwidth]{figs/sup_chexpert.pdf}
% \vspace{-3mm}
% \caption{More samples describing the qualitative performance of the DECAPS architecture on the five conditions of interest (continuing from Figure \ref{fig1}).} 
% \label{sup1}
% \end{figure}

% \begin{figure}[!t]
% \centering
% \includegraphics[width=\textwidth]{figs/sup_chexpert_2.pdf}
% \vspace{-3mm}
% \caption{Qualitative performance of DECAPS on other conditions described by the CheXpert dataset, namely (a): lung lesion, (b) pneumonia, and (c) support device. Each column of 4 images presents the predicted bounding box on the input image, the class activation map, the cropped ROI, and the second level class activation map, in that order.} 
% \label{sup1}
% \end{figure}

% \pagebreak

\vspace{5mm}
\noindent
\textbf{Acknowledgments.}
This research was supported by the National Science Foundation (1910973).

{\small
\bibliographystyle{splncs04}
\bibliography{references.bib}
}

\pagebreak

\textbf{Supplementary Material}

\begin{figure}[!h]
\centering
\includegraphics[width=0.8\textwidth]{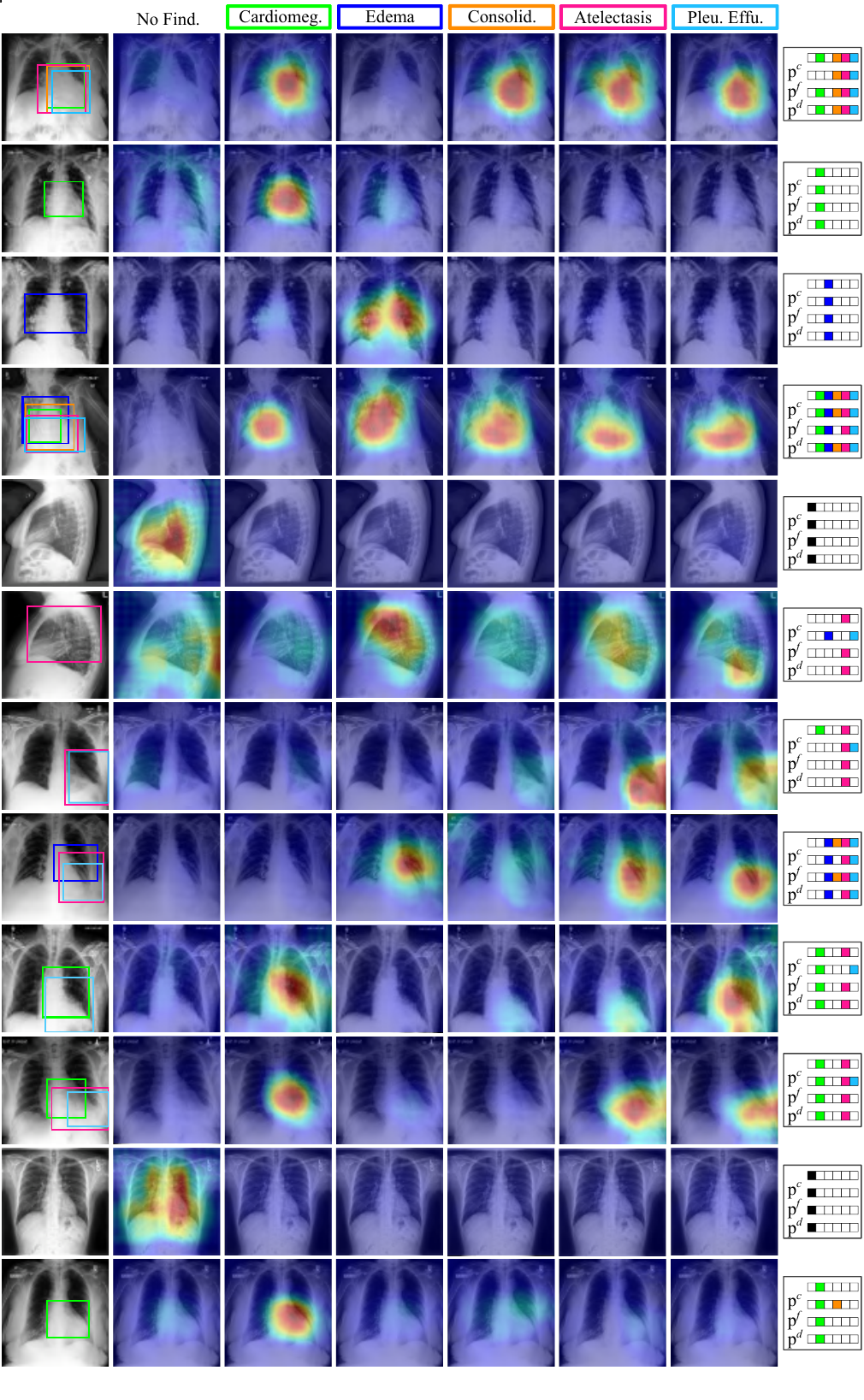}
% \vspace{-3mm}
\caption{More samples describing the qualitative performance of the DECAPS architecture on the five conditions of interest (continuing from Figure 2).} 
\label{sup1}
\end{figure}

\begin{table}
\centering
\caption{Mean AUC (\%) over the 5 selected pathologies of the CheXpert dataset to show the effect of proposed components and their combinations.}
\resizebox{0.7\columnwidth}{!}{
\begin{tabular}{c|c|c|c|c}
\hline
\quad IDR \;\;\quad & \;Patch Drop\;  & \;Patch Crop\; & \;Distillation\; & \;mAUC (\%)\; \\ \hline
 &  &  &  & \;87.24\; \\
\checkmark &  &  &  & \;89.90\; \\ \hline
\checkmark & \checkmark &  &   & \;90.57\; \\
\checkmark &  & \checkmark &   & \;91.06\; \\
\checkmark & \checkmark & \checkmark &   & \;91.56\; \\ \hline
\checkmark & \checkmark & \checkmark & \checkmark  & \;92.82\; \\ \hline
\end{tabular}
}
% \vspace{-8mm}
\label{table:2}
\end{table}

\begin{figure}[!b]
\centering
\includegraphics[width=0.9\textwidth]{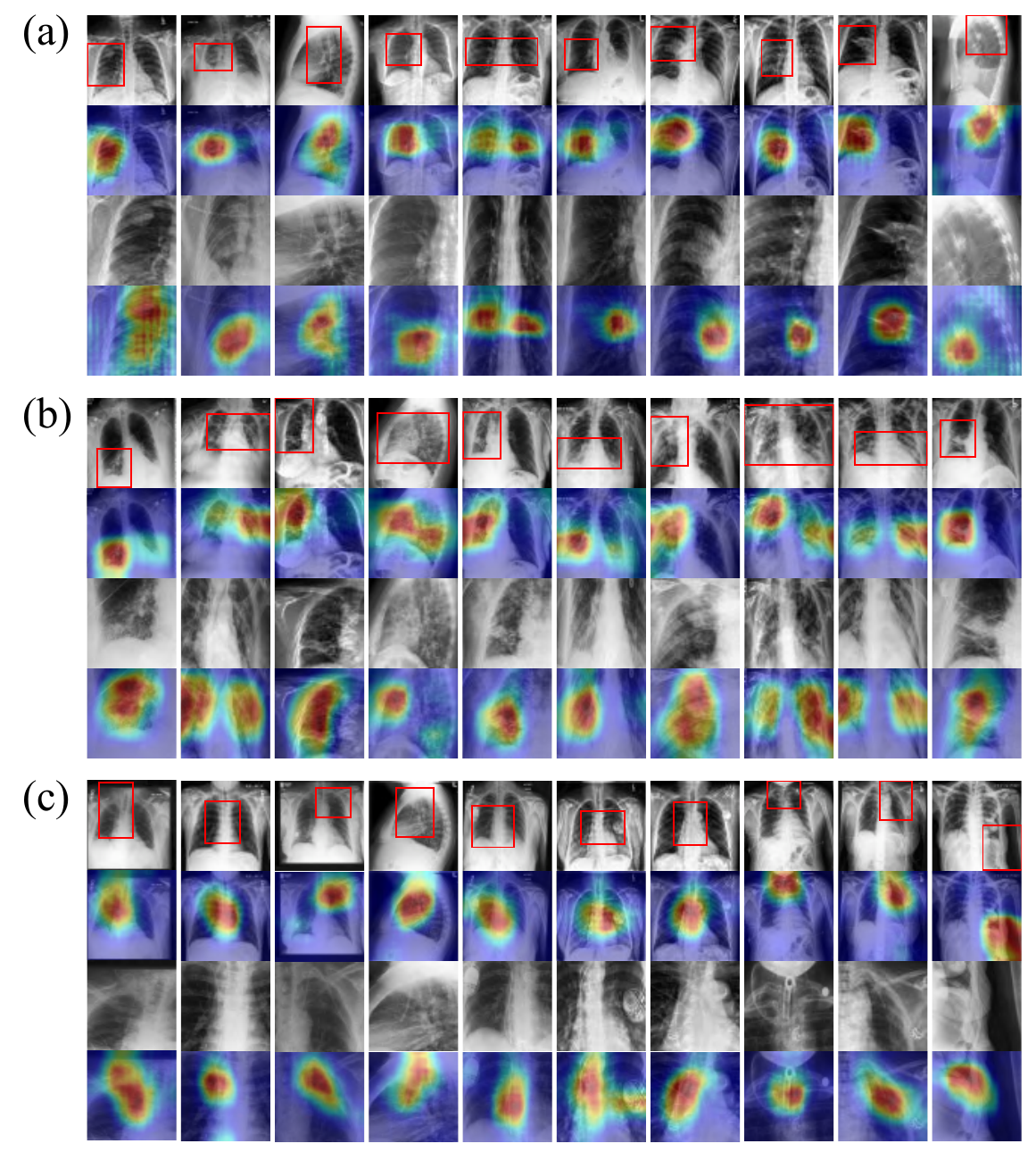}
% \vspace{-3mm}
\caption{Qualitative performance of DECAPS on other conditions described by the CheXpert dataset, namely (a): lung lesion, (b) pneumonia, and (c) support device. 
Each column of 4 images presents the predicted bounding box on the input image, the class activation map, the cropped ROI, and the second level class activation map, in that order.
} 
\label{sup1}
\end{figure}






% LNCS Homepage, \url{http://www.springer.com/lncs}. Last accessed 4
% \end{thebibliography}
\end{document}